%% file: eccv2022submission.tex
\documentclass[runningheads]{llncs}
\usepackage{graphicx}

\usepackage{tikz}
\usepackage{comment}
\usepackage{amsmath,amssymb} 
\usepackage{color}

\usepackage[accsupp]{axessibility}  

\usepackage{tikz}
\usepackage{comment}
\usepackage{amsmath}
\usepackage{amssymb}
\usepackage{color}
\usepackage{cite}
\usepackage{soul}

\usepackage{url}
\usepackage{blindtext}
\usepackage{hyperref}

\usepackage{url}
\usepackage{booktabs, multirow} 
\usepackage{soul}
\usepackage{changepage,threeparttable} 
\usepackage{float}
\usepackage{pifont}
\newcommand{\cmark}{\color{green}\checkmark}%
\newcommand{\xmark}{\color{red}\ding{55}}%
\usepackage{xcolor}
\usepackage{makecell}

\usepackage{graphicx}
\usepackage{float}
\usepackage{subfig}

\usepackage{caption}
\begin{document}
\pagestyle{headings}
\mainmatter
\def\ECCVSubNumber{807}  

\title{Efficient One Pass Self-distillation with Zipf's Label Smoothing} 

\titlerunning{Zipf's LS}
%
\author{Jiajun Liang$^\dagger$ \and Linze Li \and Zhaodong Bing \and Borui Zhao \and Yao Tang \and Bo Lin \and Haoqiang Fan}
\authorrunning{Jiajun Liang et al.}
\institute{MEGVII Technology
\email{\{liangjiajun,lilinze,bingzhaodong,zhaoborui,tangyao02,linbo,fhq\}@megvii.com}}
\maketitle

\begin{abstract}
\input{abstract}
\end{abstract}

\section{Introduction}

\input{introduction}

\section{Related Work}
\input{related_work}

\section{Method}

\input{method}

\section{Experiment}
\input{experiment}

\section{Discussion}

\input{discussion}

\section{Conclusion}
\input{conclusion}

\par\vfill\par
\clearpage
%
%
\bibliographystyle{splncs04}
\bibliography{egbib}
\end{document}


\definecolor{commentcolor}{RGB}{3, 148, 252}   
\newcommand{\PyComment}[1]{\ttfamily\textcolor{commentcolor}{\# #1}}  
\newcommand{\PyCode}[1]{\ttfamily\textcolor{black}{#1}} 
\pagestyle{headings}
\mainmatter
\def\ECCVSubNumber{807}  

\title{Supplementary Materials of \\Efficient One Pass Self-distillation with Zipf's Label Smoothing} 

\titlerunning{Zipf's LS}
%

\titlerunning{Zipf's LS}
%
\author{Jiajun Liang \and Linze Li \and Zhaodong Bing \and Borui Zhao \and Yao Tang \and Bo Lin \and Haoqiang Fan}
\authorrunning{Jiajun Liang et al.}
\institute{MEGVII Technology
\email{\{liangjiajun,lilinze,bingzhaodong,zhaoborui,tangyao02,linbo,fhq\}@megvii.com}}
\maketitle

\section{Explanation to empirical observation}

We find that the Zipf's prior could help
generate non-uniform supervision for non-target classes
in a one-pass way. In this section, we provide a simple intuition to explain why Zipf's law should occur for predictions from multi-class classification.

We postulate that one main source of the non-zero network predictions is the inevitable non-orthogonality of the inter-class feature vectors as more and more classes are packed into the finite-dimensional feature space. In a simplified model, we assume that the decision vectors corresponding to each class are uniformly distributed on a high-dimensional unit sphere. Then for another random query vector on the sphere, their inner-products with it distribute in the shape of a Gaussian when the dimension is high enough. 
\begin{figure}[btp]
  \centering
\includegraphics[width=0.4\textwidth]{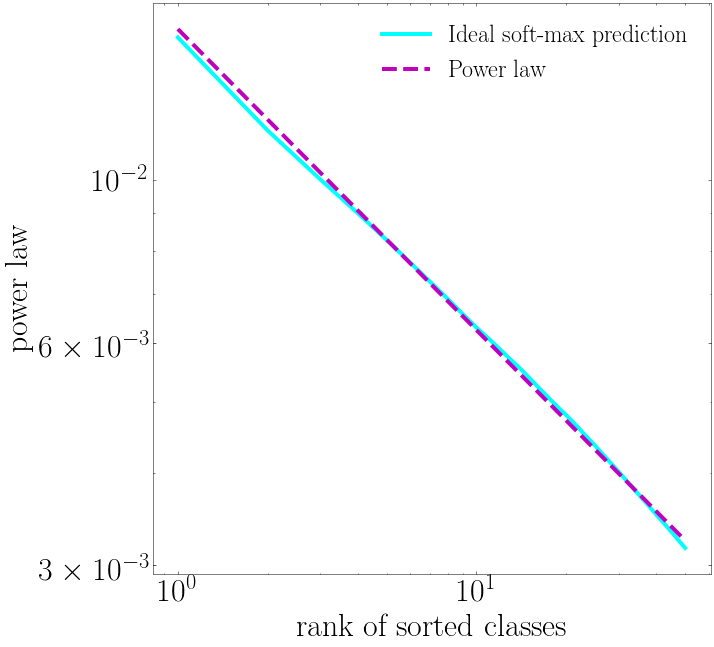}
  \caption{softmax Gaussian logits fit Zipf's law well}
  \label{fig:ideal fit}
\end{figure}

We propose a toy experiment to verify that softmax Gaussian logits fit Zipf's law well. As shown in Algorithm \ref{gaussian_logits}, first, we sampled random vectors from multivariate normal distribution $\mathcal{N}(\boldsymbol{0},\,I_{1000})$ as logits of different samples. Then logits for each sample are sorted and applied with softmax to get probabilities. At last, we average the sorted probabilities across all samples and plot the probabilities-rank relation for the top 32 categories in log-log space. It could be seen that a straight line pattern shows up in Figure \ref{fig:ideal fit}.



\begin{algorithm}[h]
\SetAlgoLined
    \PyComment{generate Gaussian logits,1000 samples, 1000 classes} \\
    \PyCode{logits = np.random.randn(1000, 1000)} \\
    \PyComment{sort in class dimension} \\
    \PyCode{sorted\_logits = np.sort(logits, axis=1)} \\
    \PyComment{probability predictions by applying soft-max on the logits} \\
    \PyCode{sorted\_preds = np.exp(sorted\_logits) / np.sum(np.exp(sorted\_logits),axis=1)[:,None]}\\ 
    \PyComment{averaged across samples} \\
    \PyCode{mean\_sorted\_preds = np.mean(sorted\_preds, axis=0)} \\
    \PyComment{top 32 ranks considered} \\
    \PyCode{top32\_sorted\_preds = mean\_sorted\_preds[::-1][:32]} \\

\caption{simulation of ranking softmax Gaussian logits}
\label{gaussian_logits}
\end{algorithm}

We also compare several most frequently-used distributions of long-tail shapes (Zipf's law, exponential and log-normal) to fit the empirical softmax scores, as shown in Table \ref{tab:laws_of_fit}. All parameters of distributions are estimated by most-likelihood estimation. Common statistical test metrics such as R square are measured. Zipf's law outperforms the other two in all kinds of metrics.

\setlength{\tabcolsep}{6pt}
\begin{table}[!htp]\centering
\caption{Statistical test of how well different distributions fit on empirical averaged predictions on INAT-21. The top 50 categories are considered. For tests such as Chisquare and Kolmogorov–Smirnov which heavily rely on the amount of the samples, we sample $10^5$ instances from the empirical distribution. $D$ is the Kolmogorov–Smirnov statistic and $p$ is the p-value. Zipf's law outperforms the other two by all kinds of metrics.}\label{tab:laws_of_fit}
\scriptsize
\begin{tabular}{lcccc}\toprule
\textbf{Metric} &\textbf{Zipf's law} &\textbf{Exponential} &\textbf{Log-normal} \\\cmidrule{1-4}
$R^2$ &0.99992 &0.6768 &0.9672 \\\cmidrule{1-4}
Kullback–Leibler divergence &0.0000667 &0.315 &0.0219 \\\cmidrule{1-4}
Jensen–Shannon divergence &0.0000167 &0.063 &0.00544 \\\cmidrule{1-4}
Chisquare &13.3 &451677 &4499 \\\cmidrule{1-4}
Kolmogorov–Smirnov &$D$=0.00278 &$D$=0.265 &$D$=0.0823 \\
 & $p$=0.42 &$p$=0.0 &$p$=0.0 \\
\bottomrule
\end{tabular}
\end{table}



\setlength{\tabcolsep}{6pt}
\begin{table}[ht]\centering
\caption{The detail of hyper parameters for different datasets.}\label{tab: hyper_params}
\scriptsize
\begin{tabular}{lccccc}\toprule
Dataset &$\lambda$ &$\alpha$ &dense layer &$\beta$ \\\cmidrule{1-5}
CIFAR100 &0.1 &1.0 &2 &0.1 \\
TinyImageNet &1.0 &1.0 &2 &0.5 \\
ImageNet &0.1 &1.0 &1 &/ \\
INAT21 &1.0 &1.0 &1 &/ \\
\bottomrule
\end{tabular}
\end{table}

\section{More Experiment Details and Discussion}

\subsection{Hyperparameters}
\textbf{Hyperparameters setting rules.} Table \ref{tab: hyper_params} shows the detail of hyperparameters settings for different tasks. $\alpha$ controls the decay shape of
Zipf’s distribution, and is set to 1.0 in all tasks. $\lambda$ controls the regularization strength, which is set to 0.1 for CIFAR100 and ImageNet, and 1.0 for TinyImageNet and NAT21. For datasets with large-resolution inputs, such as ImageNet and INAT, using the final dense feature maps would be sufficient, and no more dense layers are required. For low-resolution tasks such as CIFAR100 and TinyImageNet, we use one more dense layer to get enough votes for dense ranking. In this case, we need $\beta$ to weigh the cross-entropy loss for learning the extra classifier. $\beta$ is set to 0.1 and 0.5 respectively on CIFAR100 and TinyImageNet. \\

\noindent \textbf{Hyperparameters ablation study.} 1) $\alpha$ is not sensitive where $\alpha\in[0.5, 1.5]$. 2) $\lambda$ is to control regularization strength and is positively correlated with the train/val acc gap. For tasks that are prone to overfitting(TinyImageNet and INAT whose train/val gap are $35\%$ and $25\%$), $\lambda$ is $1.0$. For tasks that are less overfitting(ImageNet train/val gap is $4\%$), $\lambda$ is $0.1$. 3) $\beta$ is optional and only recommended for small resolution tasks. It should be less than $0.5$ to avoid shadow learning of deeper layers. See Table \ref{tab:params_ablation} for details.

\begin{table}[!htp]\centering
\caption{Ablation study of hyperparameters $\alpha$,$\lambda$ and $\beta$ }\label{tab:params_ablation}
\scriptsize
\begin{tabular}{ccccccc}\toprule
$\alpha$ & 0.1 & 0.5 & 1.0 & 1.5  & 2.0 &\\\midrule
CIFAR100 & 77.21±0.29 &77.26±0.13 & \textbf{77.38±0.32} & 77.12±0.24 & 76.45±0.12 & \\
TinyImageNet & 58.85±0.16 & 59.06±0.21 & \textbf{59.25±0.20} & 58.64±0.18 & 53.35±0.41 & \\
\midrule\midrule
$\lambda$ & 0.01 & 0.1 & 0.5 & 1.0  & 1.5 &\\\midrule
CIFAR100 & 76.59±0.15 & \textbf{77.38±0.32} & 76.62±0.24 & 76.79±0.04 & 76.92±0.17 & \\
TinyImageNet & 56.86±0.36 & 57.65±0.01 & 58.41±0.17 & \textbf{59.25±0.20} & 58.03±0.27 & \\
\midrule\midrule
$\beta$(optional) & 0.05 & 0.1 & 0.3 & 0.5  & 0.7 &\\\midrule
CIFAR100 & 77.24±0.22 & \textbf{77.38±0.32} & 76.75±0.31 & 76.39±0.16 & 76.49±0.24 & \\
TinyImageNet & 58.71±0.17 & 58.77±0.18 & 59.48±0.31 & \textbf{59.25±0.20} & 59.00±0.12 & \\
\bottomrule
\end{tabular}
\end{table}


\subsection{SNR of Ranking}
\begin{figure}[t]
    \centering
    \subfloat[SNR-Rank relation across all classes]{
    \includegraphics[width=0.4\textwidth]{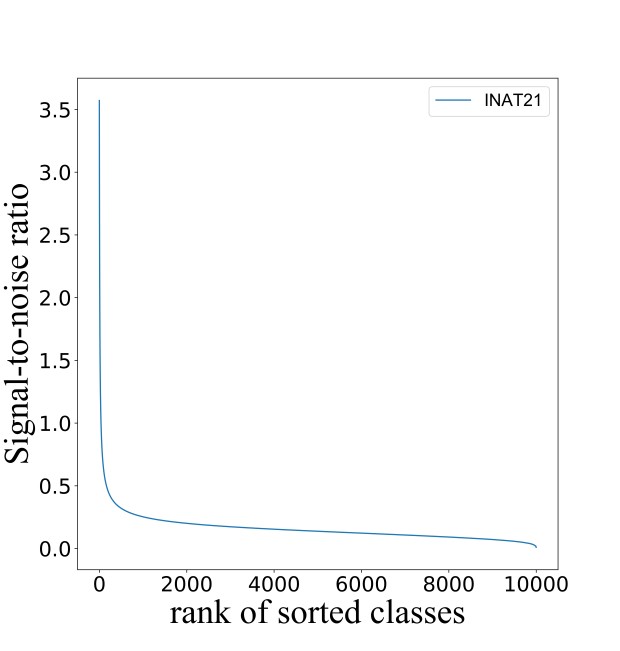}
    }\hskip2em
    \subfloat[SNR-Rank relation across top 40 classes]{
    \includegraphics[width=0.4\textwidth]{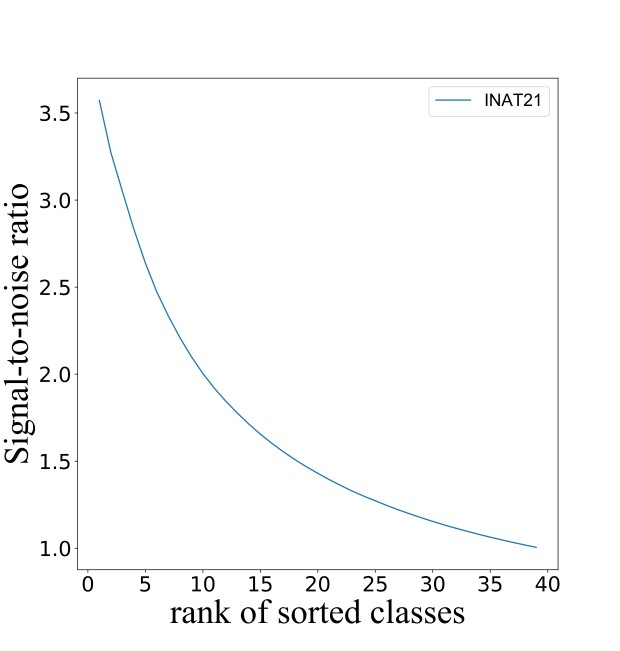}
    }
    \caption{SNR-Rank relation plot for trained ResNet-50 model in INAT21 dataset. It can be seen that only the top 40 rankings have SNR larger than 1, which makes it hard to give reliable ranks for tailing class on the fly.
    }
    \label{fig:snr}
\end{figure}

Ranking the classes accurately is a key factor to generate proper Zipf's law distribution for the sample. In the method section, we propose a finer ranking method named dense classification ranking which exploits spatial classification results from the last few feature maps. A consequence of
this voting-based method is that we have to clip the Zipf’s
values to a uniform one after a certain rank, as they would
not receive sufficient votes to be distinguished individually.
To illustrate that ranking only a few top non-target classes is sufficient, we study the signal-to-noise ratio of rankings. The signal and noise of specific rank $r$ are calculated as the average and standard deviation among $r$-th probabilities from different sorted samples respectively. As shown in Figure.\ref{fig:snr}, we plot the SNR-rank curve on the INAT21 dataset, only the top 40 out of 10000 ranks whose SNR is larger than one. It's a good trade-off to just give power-law decayed probabilities to the top-ranking class since the SNR of tailing ranks is too low to give reliable ranks. 

\subsection{More Zipf's Soft Label Examples}
\begin{figure}[hbp]
  \centering
\includegraphics[width=1.0\textwidth]{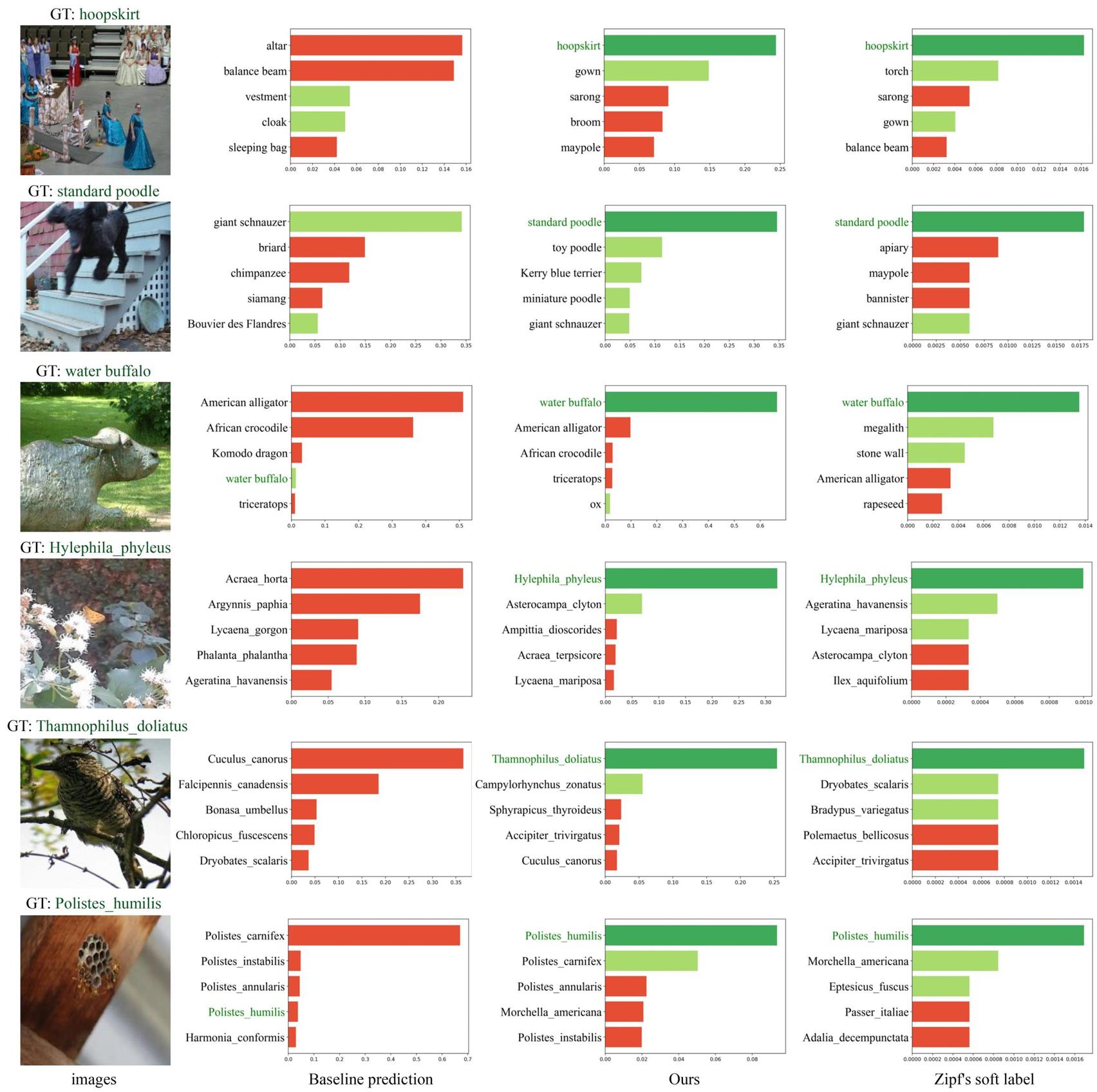}
  \caption{Top-5 predictions visualization of our proposed method (Zipf's label smoothing) compared with the baseline method (cross-entropy). 
  The dark green, light green and red denote ground-truth, similar and irrelevant categories respectively.
  ``GT'' denotes the ground truth label and thus the hard label of the baseline method. 
  The baseline prediction is acquired under the supervision of the hard label and misclassified on the samples. 
  Our method exploits target-relevant categories to better represent the image, and obtains better results.
  }
  \label{fig:result_more}
\end{figure}

Figure~\ref{fig:result_more} illustrates more results of the top-5 predictions of our proposed method compared with the baseline method. The top three rows are sampled from ImageNet while the bottom three rows are sampled from INAT21.
It can be seen that: 
(1) There are several categories similar to the target class shown up in Zipf's soft labels, which provide meaningful label representations for the network to better grasp the concept of the target class.
(2) Fine-grained categories of the target class emerge in Zipf's soft labels, which can provide similar ``dark knowledge'' as knowledge distillation. \\

\subsection{Generalization: Performance on downstream tasks}
\begin{figure*}[t]
  \centering
\includegraphics[width=1.\textwidth]{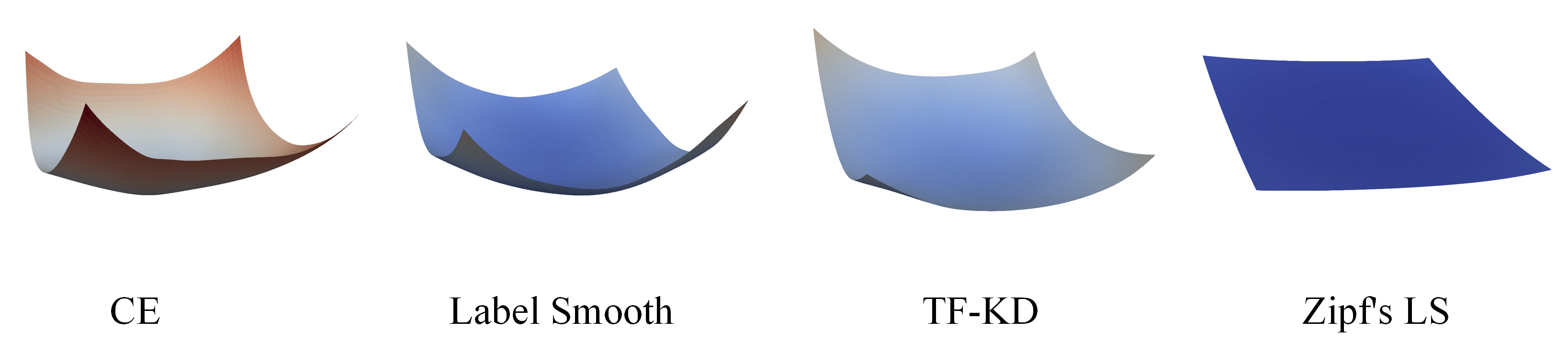}
  \caption{Comparison of loss landscape with efficient teacher-free methods}
  \label{fig:landscape}
\end{figure*}

To measure the power of generalization of Zipf's LS, we conducted the transfer learning task by fine-tuning ImageNet pre-trained models on MS-COCO, as shown in Table \ref{tab:COCO}. Besides, we visualize loss landscapes\cite{li2018visualizing} of several efficient teacher-free methods(see Fig \ref{fig:landscape}), Zipf's LS achieves more flat convergence, which is a possible hint for better generalization\cite{minima3}.

\begin{table}[!htp]\centering
\caption{ImageNet pretrained ResNet50 for object detection}\label{tab:COCO}
\scriptsize
\begin{tabular}{lccccc}\toprule
Method & Vanilla(CE) &TF-KD & PS-KD& Zipf's LS(Ours)  & \\\midrule
AP &36.4\% & 36.4\%  & 36.5\%  &\textbf{36.6\%}\\
AP@0.5 &58.3\%&  56.7\% & 56.7\%&\textbf{58.8\%} \\
\bottomrule
\end{tabular}
\end{table}

\par\vfill\par
\clearpage
%
%
\bibliographystyle{splncs04}
\bibliography{egbib}

%% file: abstract.tex
Self-distillation exploits non-uniform soft supervision from itself during training and improves performance without any runtime cost. However, the overhead during training is often overlooked, and yet reducing time and memory overhead during training is increasingly important in the giant models' era. This paper proposes an efficient self-distillation method named Zipf's Label Smoothing (Zipf's LS), which uses the on-the-fly prediction of a network to generate soft supervision that conforms to Zipf distribution without using any contrastive samples or auxiliary parameters. Our idea comes from an empirical observation that when the network is duly trained the output values of a network's final softmax layer, after sorting by the magnitude and averaged across samples, should follow a distribution reminiscent to Zipf's Law in the word frequency statistics of natural languages. 
By enforcing this property on the sample level and throughout the whole training period, we find that the prediction accuracy can be greatly improved. Using ResNet50 on the INAT21 fine-grained classification dataset, our technique achieves +3.61\% accuracy gain compared to the vanilla baseline, and 0.88\% more gain against the previous label smoothing or self-distillation strategies. The implementation is publicly available at \url{https://github.com/megvii-research/zipfls}.


%% file: introduction.tex

\begin{figure}[t]
  \centering
\includegraphics[width=1.0\textwidth]{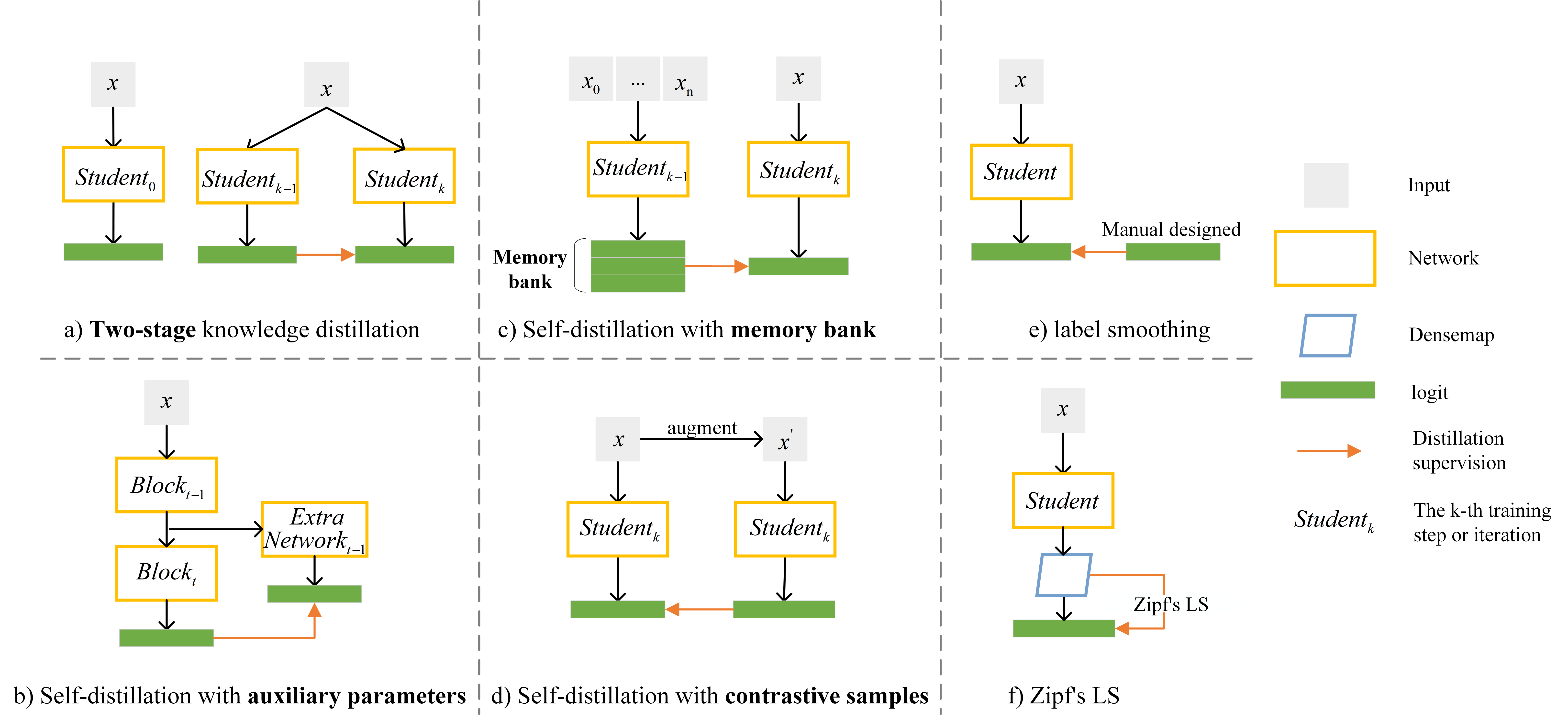}
  \caption{Comparison between different soft label generation methods. a) \textbf{Two-stage} knowledge distillation method~\cite{furlanello2018born, hinton2015distilling}. b) \textbf{Auxiliary parameters}~\cite{zhang2019your, guo2020online, ji2021refine}. c) Progressive distillation with \textbf{memory bank} storing past predictions of the entire dataset ~\cite{kim2021self, bagherinezhad2018label, zhang2021delving}. d) Contrastive samples~\cite{xu2019data, yun2020regularizing} with \textbf{twice training iterations}. e) Label smoothing methods~\cite{muller2019does, yuan2020revisiting}, with \textbf{uninformative} manual designed distributions. f) \textbf{Efficient one-pass} Zipf's Label Smoothing method, generating sample-level non-uniform soft labels almost without additional cost during training}
  \label{fig:intro_compare}
\end{figure}

A major trend in the study of multi-class classification models is to replace the one-hot encoding label with more informative supervision signal. This line of thought witnessed proliferation of great training techniques to improve the accuracy of a network without any runtime cost, the Knowledge Distillation being the most famous of them. Perhaps the most counter-intuitive discovery in this direction is the effectiveness of Self Distillation methods~\cite{furlanello2018born,yuan2020revisiting,xu2019data,yun2020regularizing, Zhang2021SelfDistillationTE} where a network even benefits from predictions of its own.

Self distillation simplifies the two-stage knowledge distillation framework by distilling knowledge from itself instead of from the pretrained teacher, and still improves performance significantly without extra cost in inference time. However, the overhead during training in self-distillation is often overlooked, and yet reducing time and memory overhead in training is increasingly important in today's giant model era. Fig. \ref{fig:intro_compare} shows several knowledge distillation paradigms, and self distillation methods rely on additional contrastive training instances, auxiliary parameters or intermediate dumped results for each training sample, which could double the training time and bring non-negligible memory overhead.

\begin{figure}[t]
    \centering
    \subfloat[different network architectures]{
    \includegraphics[width=0.45\textwidth, height=5cm]{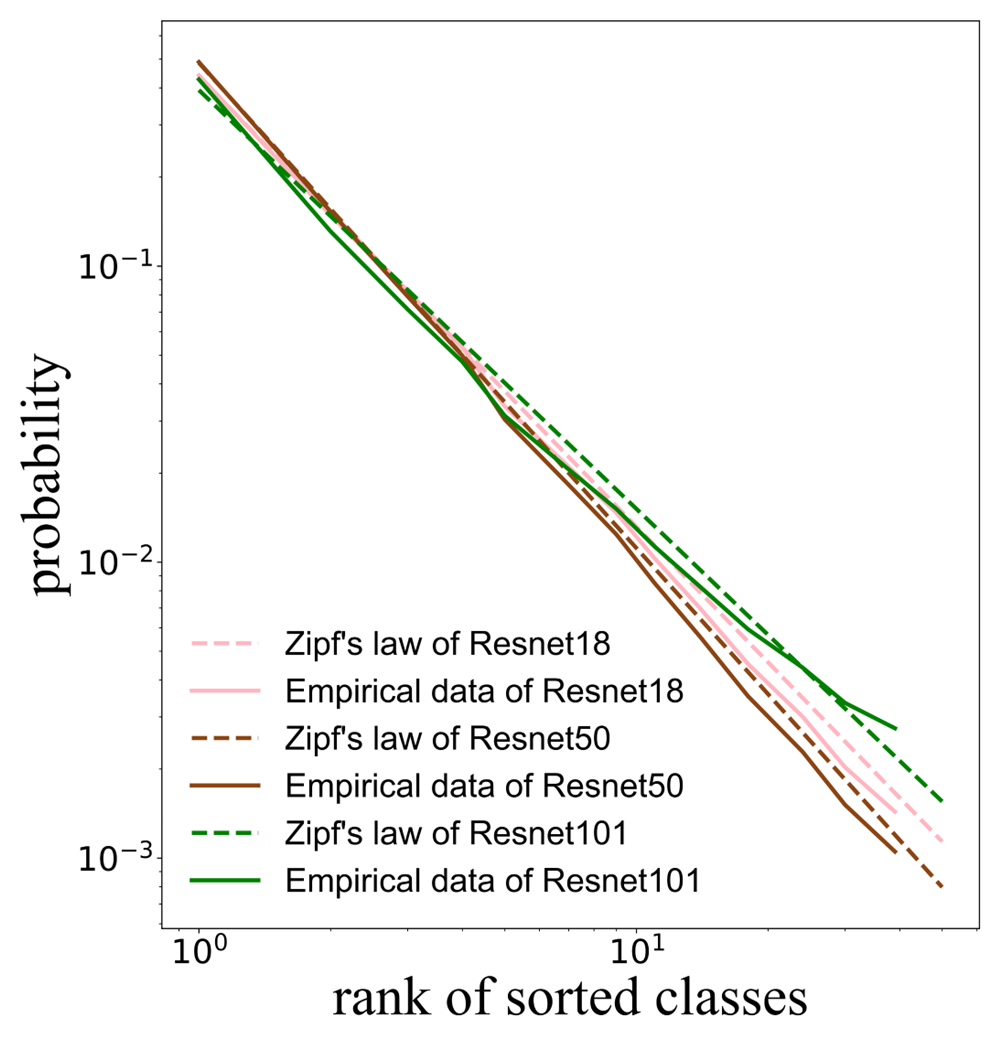}
    }
    \subfloat[different datasets]{
    \includegraphics[width=0.45\textwidth, height=5cm]{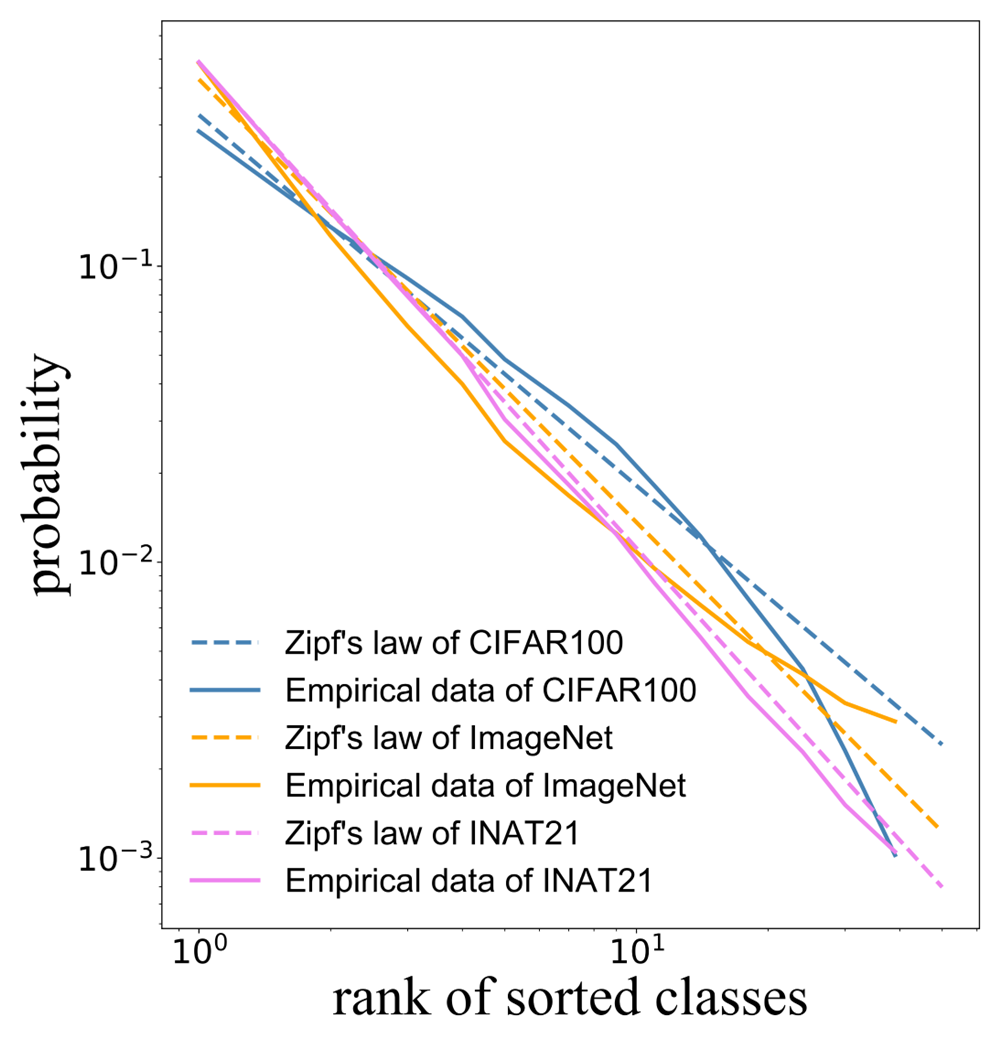}
    }
    \caption{ Distribution of sorted softmax scores from duly-trained networks of a)different architectures on INAT-21 and b) ResNet-50  on different datasets. The average distribution of sorted softmax values (the solid lines) well follows Zipf's law (the dashed lines of the same color). The probability-rank relations form straight lines in the log-log plot. This inspires us to design the soft supervision similar to this shape
    }
    \label{fig:prediction_zipf_fit}
\end{figure}

This paper aims to find efficient techniques which generate non-uniform supervision signals as informative as expensive self distillation approaches. The starting point of our construction is the observation of the network's soft-max output values. The class that corresponds to the final categorical output usually has the highest value, but the scores for other classes (which we call the non-target classes), 
containing important information of the network's understanding of the input image, also play an important role. Indeed,  we surmise that \textbf{a good-performing network should obey certain laws in the non-zero prediction values they make on the non-target classes}.

We postulate that a significant part of the efficacy of the distillation techniques comes from enforcing the prediction scores into a shape that best balances between being ``sharp'' (so that the final prediction is unambiguous) and being ``soft'' (so that inter-class correlation are respected). We test this hypothesis by inventing a technique, which only uses the on-the-fly prediction of a network to generate a soft supervision label that conforms to our designated distribution law, and showing that this simple strategy already harvests or even surpasses the performance gains of many other more complex distillation techniques.

Specifically, we rank the output classes according to the feature output of the multi-class classification network, and assign to each class a target value according to Zipf's Law Distribution:
$$
\boldsymbol{p} \propto \boldsymbol{r}^{-\alpha}
$$
where $\boldsymbol{p}$ is the confidence distribution of different classes, $\boldsymbol{r}$ is our sorted rank index of classes (integer values starting from 1), and $\alpha$ is a hyper-parameter that controls the rate of decay. Divergence to this smoothed label is added as a loss term in complement to the usual cross-entropy with the one-hot encoded hard label.

The choice of Zipf's Law is not arbitrary. Indeed, we experimentally find out that when a network is trained to its convergence state, the average distribution of sorted softmax values well follows this law (see Fig.~\ref{fig:prediction_zipf_fit}). By explicitly enforcing this shape of the distribution and giving supervision from the very beginning of training, the performance of the network boots by a significant amount.

\begin{table}[!htp]\centering
\caption{Comparison between different soft label generation methods. Our Zipf's Label Smoothing generates sample-level non-uniform soft labels with little cost during training. For simplicity, this table only shows top1-accuracy(Acc) from ResNet-18 on TinyImageNet(Tiny) and ResNet-50 on ImageNet(IMT). More comprehensive results on other models and datasets are reported in the remaining sections of this paper. Memory cost and training time test experiments are conducted on ImageNet using 4 2080Ti GPUs with batch size 16}\label{tab: method_compare }
%
\scriptsize
\resizebox{\textwidth}{2cm}{
\begin{tabular}{lccccccccc}\toprule
Method &\makecell{Non-\\Uniform} & \makecell{w/o\\ Pretrain \\ Teacher} & \makecell{w/o\\ Contrastive\\Samples} & \makecell{w/o \\ Auxiliary \\ Parameters} & \makecell{CPU\\ Memory \\Cost} &\makecell{GPU \\Memory \\Cost} &\makecell{Training \\time per \\epoch} & \makecell{Tiny \\Acc \\(gain)}  & \makecell{IMT \\Acc \\(gain)} \\\midrule
\color{black}Baseline  &\xmark &\cmark &\cmark &\cmark &\color{black}18G &\color{black}6.8G &\color{black}1.82h &\color{black}56.41 &\color{black}76.48 \\
\color{black}BAN \cite{furlanello2018born} &\cmark  &\xmark &\cmark &\xmark & \color{black}18.9G &\color{black}7.4G &\color{black}2.67h &\color{black}(+2.24) &\color{black}(+0.05)\\
\color{black}BYOT \cite{zhang2019your}  &\cmark  &\cmark &\cmark &\xmark &\color{black}18.2G &\color{black}38.5G &\color{black}10.36h &\color{black}(+1.43) &\color{black}(+0.51)\\ 
\color{black}PS-KD \cite{kim2021self}  &\cmark &\cmark &\xmark &\cmark  &\color{black}27.4G &\color{black}9.8G &\color{black}2.18h &\color{black}(+1.81) &\color{black}(+0.18)\\
\color{black}DDGSD \cite{xu2019data}  &\cmark &\cmark &\xmark &\cmark & \color{black}18.2G &\color{black}7.3G &\color{black}4.30h &\color{black}(+2.11) &\color{black}(+0.46)\\
\color{black}CS-KD \cite{yun2020regularizing} &\cmark &\cmark &\xmark &\cmark &\color{black}18G &\color{black}6.8G &\color{black}2.86h &\color{black}(+1.97) &\color{black}(+0.30)\\
\color{black}LS \cite{szegedy2016rethinking}  &\xmark  &\cmark &\cmark &\cmark &\color{black}18G &\color{black}6.8G &\color{black}1.83h &\color{black}(+0.48) &\color{black}(+0.19)\\
\color{black}TF-KD \cite{yuan2020revisiting}  &\xmark  &\cmark &\cmark &\cmark &\color{black}18G &\color{black}6.8G &\color{black}1.82h &\color{black}(+0.26) &\color{black}(+0.08)\\ \midrule
\color{black}Ours &\cmark &\cmark &\cmark  &\cmark &\color{black}18G &\color{black}6.8G &\color{black}1.83h & \color{black}\textbf{(+2.84)}  &\color{black}\textbf{(+0.77)} \\
\bottomrule

\end{tabular}}
\end{table}

Thus we propose an efficient and general plug-and-play technique for self distillation, named Zipf's Label Smoothing (Zipf's LS). Compared to other techniques (see Table~\ref{tab: method_compare }), our method enjoys the advantage of incurring almost zero additional cost (during inference or training) as label smoothing and at the same time strongly preserves the performance gains of the self-distillation.
To summarize, our contributions are as follows:

\begin{itemize}
  \item We find that distribution of non-target soft-max values of duly trained model fits well to Zipf's law, which could be used as a regularization criterion in the entire self-training process.
  \item We propose Zipf's Label Smoothing method, an efficient self distillation training technique without relying on additional contrastive training instances or auxiliary parameters.
  \item We verify our method on comprehensive combinations of models and datasets (including popular ResNet and DenseNet models, CIFAR, ImageNet and INAT classification tasks) and show strong results.
\end{itemize}

%% file: related_work.tex
\subsection{Label Smoothing}
The one-hot label is sub-optimal because objects from more than one class occur in the same image.
Label Smoothing~\cite{szegedy2016rethinking} (LS) is proposed to smooth the hard label to prevent over-confident prediction and improve classification performance.
M{\"u}ller et al.~\cite{muller2019does} found that intra-class distance in feature space is more compact when LS is used, which improves generalization.
To obtain non-uniform soft labels, Zhang et al.~\cite{zhang2021delving} proposed the online label smoothing method (OLS) by maintaining the historical predictions to obtain the class-wise soft label. Yuan et al.~\cite{yuan2020revisiting} discussed the relationship between LS and knowledge distillation, and proposed a teacher-free knowledge distillation (Tf-KD) method to get better performance than LS. Label smoothing has become one of the best practices in the current deep learning community~\cite{he2019bag}, but the paradigm of using uniform distribution for the non-target classes limits further improvement of performance. 

\subsection{Knowledge Distillation}
Instead of imposing a fixed prior distribution, knowledge distillation was first proposed by Hinton in~\cite{hinton2015distilling} to provide sample-level non-uniform soft labels. They demonstrated that the ``dark knowledge'' lies in the output distributions from a large capacity teacher network and benefits the student's representation learning. 
Recent works mainly explored to better transfer the ``dark knowledge'' and improve the efficiency from various aspects, such as reducing the difference between the teacher and student~\cite{cho2019efficacy, mirzadeh2020improved, zhu2021student, beyer2021knowledge}, designing student-friendly architecture~\cite{park2021learning, Ma2021UndistillableMA}, improving the distillation efficiency~\cite{kim2021self, furlanello2018born, xu2019data, yun2020regularizing} and explaining the distillation's working mechanism~\cite{stanton2021does, allen2020towards}.

In this work, we focus on how to transfer the ``dark knowledge'' in an almost free manner.
Furlanello et al.~\cite{furlanello2018born} proposed to improve the performance of the student network by distilling a teacher network with the same architecture.
However, it is still a two-stage approach, which first trains the teacher and then distills knowledge to the student. To reduce the training time, many self-distillation methods were proposed. They gain soft label supervision on the fly without the pretraining step.

\subsection{Self Distillation}
There are two categories of self-distillation techniques, namely the auxiliary parameter methods~\cite{guo2020online, zhang2019your, bagherinezhad2018label, ji2021refine, zagoruyko2016paying} and contrastive sample  methods~\cite{xu2019data, yun2020regularizing, zhang2021delving, kim2021self}. Auxiliary model methods exploit additional branches to get extra predictions besides the main-branch prediction for soft label supervision at the cost of more parameters overhead. For example, Knowledge Distillation via Collaborative Learning (KDCL)~\cite{guo2020online} trained multiple parallel student networks at the same time and ensemble the output as extra soft label supervision for each parallel student network. On the other hand, contrastive sample methods get soft label supervision at the cost of additional data augmentation, enlarged batch size, or complex sampling strategy. The examples are Data-distortion Guided Self-Distillation (DDGSD)~\cite{xu2019data} which gains soft labels from different augmented views from the same instance and Regularizing Class-wise Predictions via Self-knowledge Distillation (CS-KD)~\cite{yun2020regularizing} which gathers data from other samples of the same class.

As summarized above, Label Smoothing and Knowledge Distillation are two major techniques to acquire informative soft labels. However, the Label Smoothing methods are limited by the uniform hypothesis, while the Knowledge Distillation methods require much more memory or computation overhead. Our work aims to improve upon these issues.

%% file: method.tex
\begin{figure*}[t]
  \centering
\includegraphics[width=1.\textwidth]{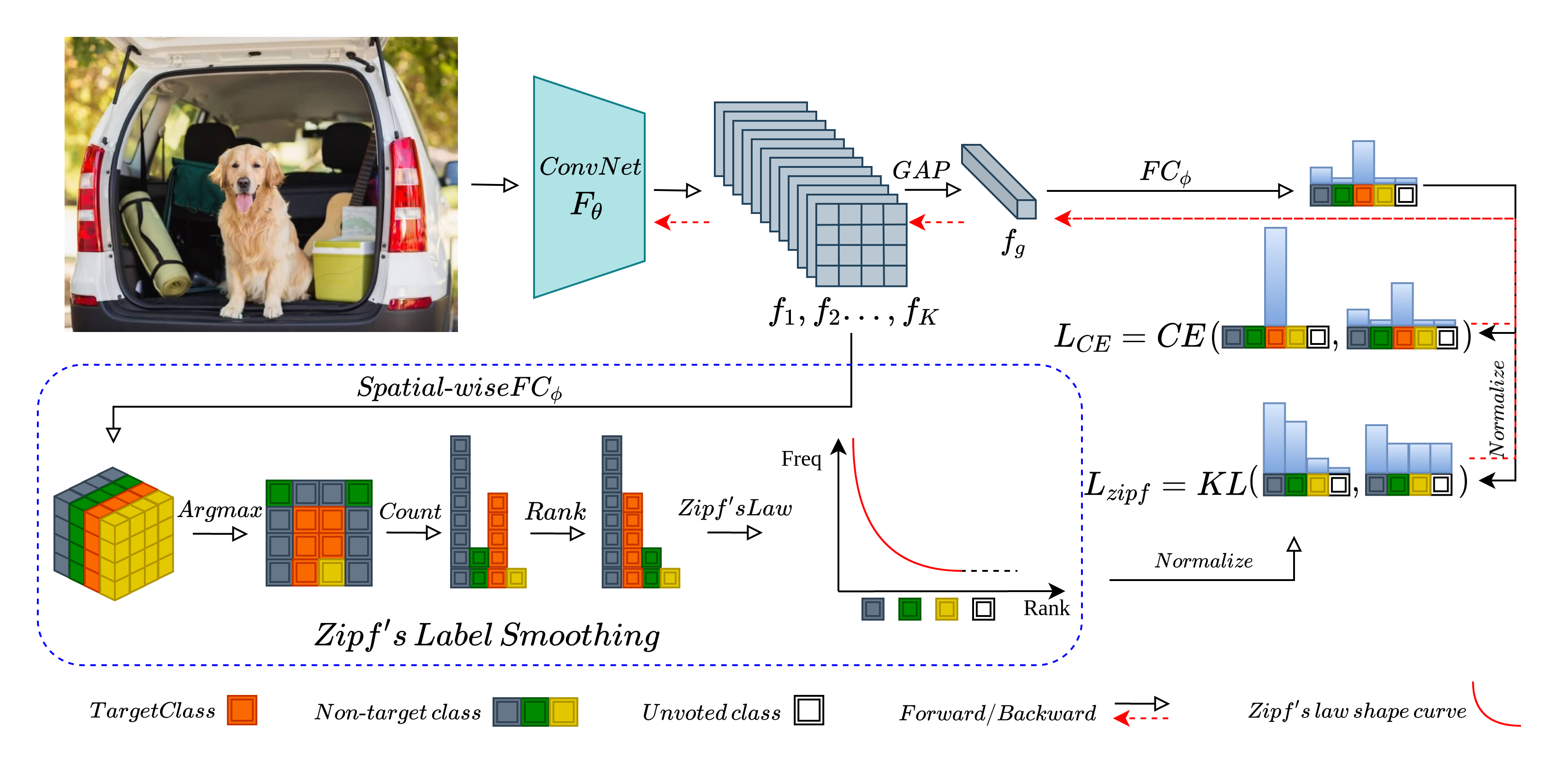}

  \caption{The overall Zipf's soft label generation and training framework. The blue dashed box outlines the soft label generation process. We apply a shared-classifier on dense feature maps and count the number of argmax values from the dense prediction, which could provide ranking information to Zipf's law distribution generation. For the non-occurred classes with zero count, we give uniform constant energy to them. $L_{zipf}$ is Kullback–Leibler divergence between the prediction of non-target classes and zipf's soft label, combined with $L_{CE}$ of the hard label to provide gradient to representation learning}
  \label{fig:overall framework}
\end{figure*}

Zipf's Label Smoothing aims to combine the best of the two worlds, the efficiency of teacher-free label smoothing and the informative soft label from self distillation. It generates non-uniform supervision signals from on-the-fly prediction of the network as shown in Fig. \ref{fig:overall framework}. Our method is inspired by the observation that the value and rank of the softmax output from the duly trained network  follow a distribution reminiscent to the Zipf's Law on average as shown in Fig. \ref{fig:prediction_zipf_fit}, which could be applied as a shape prior to the softmax prediction during the whole training period. To apply Zipf's Law to soft-label generation, ranking information of output categories is needed. We propose Dense Classification Ranking which utilizes local classification results to rank the categories. Finally, KL-divergence between prediction and Zipf's soft label within non-target classes is measured to provide more informative gradients for representation learning.


\subsection{Zipf's Law}
Zipf's law is an empirical law which states that the normalized frequency of an element should be inversely proportional to the rank of the element, firstly discovered by G.Zipf on linguistic materials \cite{powers1998applications}. It can be described by an equation as:
\begin{equation} \label{zipf equation}
\begin{split}
    f(r) &= \frac{r^{-\alpha}}{\sum_{r=1}^{N} r^{-\alpha}}\\
    log f(r) &= -\alpha \log r - \log (\sum_{r=1}^{N} r^{-\alpha})\\
\end{split}
\end{equation}
where $r$ is the rank of the element, $N$ is the total number of elements, $f$ is the frequency and $\alpha$ is a constant larger than zero which controls the decay rate.

An interesting discovery is that the outputs of softmax networks follow Zipf's Law~\cite{powers1998applications} when the network is trained to its convergence state, and this pattern consistently emerges in different datasets and models as shown in Fig.~\ref{fig:prediction_zipf_fit}. We could exploit this shape prior to design a simple and efficient self-distillation algorithm. To generate a soft label using Zipf's distribution, ranking information of categories is a must which is almost impossible to get from annotation.

\subsection{Dense Classification Ranking}\label{spatial-dense vote rank}

To realize the Zipf distribution we need to find a way to properly rank the output categories. The naive thought is to directly sort the softmax prediction of a sample, which we call the logit-based ranking method. Though this method already generates performance gain (as shown in section~\ref{section:ablation}), we find a finer treatment of the relative ranking of the top classes to be beneficial.

Common image classification convolution networks extract a feature map $\boldsymbol{F}$ from an image sample. A global average pooling ($GAP$) is then applied to $\boldsymbol{F}$.
The fully-connected layer ${FC}$ and soft-max operation will output the logits $\boldsymbol{z}$ and the final prediction $\boldsymbol{p}$.

If we used ${FC}$ directly on every pixel of the dense feature map $\boldsymbol{F}$, we cat get local classification results  $\boldsymbol{p}^{L}$:

\begin{equation} \label{local_eq}
\begin{split}
\boldsymbol{p}_k^{L} &= Softmax(FC_\phi(\boldsymbol{F}_k)), k=1,2,...,H\times W\\
\end{split}
\end{equation}
\par
where $L$ is just a mark to show the difference between global prediction $\boldsymbol{p}$ and local predictions $\boldsymbol{p}^L$. These predictions give a more complete description of what the image contains, as the object of the target class usually only occupies part of the image. We take this information into account by identifying the individual top-1 class from each location as a vote and aggregate the votes into a histogram. The classes are finally ranked by their frequency of appearance in the histogram. The remaining classes will share the same lowest rank.

\subsection{Zipf's Loss}
Zipf's loss $L_{Zipf}$ of image sample $\boldsymbol{x}$ and label $y$ is defined as KL-divergence $D_{KL}$ between normalized non-target prediction $ \boldsymbol{\hat{p}} $ and the synthetic Zipf's soft label $ \boldsymbol{\tilde{p}}$ generated from dense classification ranking:

\begin{equation} \label{eq_KL_zipf}
\begin{split}
\boldsymbol{\hat{p}}_c&=\frac{\exp{(\boldsymbol{z}_{c})}}{\sum_{m=1,m\neq y}^{C}{\exp{(\boldsymbol{z}_{m})}}}\\
L_{Zipf}(\boldsymbol{x},y) &= D_{KL}(\boldsymbol{\tilde{p}}||\boldsymbol{\hat{p}}) =\sum_{c=1,c\neq y}^{C} \boldsymbol{\tilde{p}}_c log \frac{\boldsymbol{\tilde{p}}_c}{\boldsymbol{\hat{p}}_c} \\
\end{split}
\end{equation}
\par

The synthetic Zipf's label $\boldsymbol{\tilde{p}}$ of non-target class $c$ should follow the equation \ref{prob_zipf} of Zipf's law with the corresponding rank $\boldsymbol{r}_c$:


\begin{equation} \label{prob_zipf}
\begin{split} 
    \boldsymbol{\tilde{p}} _c&= \frac{{\boldsymbol{r}_{c}}^{-\alpha}}{\sum_{m=1,m\neq y}^{C} {\boldsymbol{r}_{m}}^{-\alpha}} \\
\end{split}
\end{equation}
\par

where $\alpha$ is a hyper-parameter that controls the shape of the distribution.

The $L_{CE}$ is the standard cross entropy loss with one-hot ground-truth label. The combined loss function is as follow:
\begin{equation} \label{eq_KL_totalloss}
\begin{split}
Loss(\boldsymbol{x},y) =L_{CE}(\boldsymbol{x},y)+\lambda L_{Zipf}(\boldsymbol{x},y)
\end{split}
\end{equation}
$L_{CE}$ encourages the prediction to be sharp and confident in the target class, while $L_{Zipf}$ regularized the prediction to be soft within non-target classes.$\lambda$ is a hyper-parameter which control the regularization strength.

\noindent \textbf{Comparison with Uniform Label Smoothing.}
The gradient with respect to non-target logits for $L_{zipf}$ and ${L_{LS}}$ are shown as:

\begin{equation}\label{gradient}
\begin{split}
\frac{\partial L_{Zipf}(\boldsymbol{x},y)}{\partial \boldsymbol{z}_{c}}
 &= \begin{cases}
0 &\text{$c = y$}\\
\boldsymbol{\hat{p}}_c-\boldsymbol{\tilde{p}}_c &\text{$c \neq y$}
\end{cases} \\
\frac{\partial L_{LS}(\boldsymbol{x}, y)}{\partial \boldsymbol{z}_{c}}
 &= \begin{cases}
\boldsymbol{p}_c - (1-\beta) &\text{$c = y$}\\
\boldsymbol{p}_c - \frac{\beta}{C-1} &\text{$c \neq y$}
\end{cases} \\
\end{split}
\end{equation}

Label smoothing generates a soft label with a uniform value $\frac{\beta}{C-1}$ for all non-target classes and $1-\beta$ for the target class, where $C$ is the total number of classes. However, label smoothing suppresses predictions of high-ranked classes or promotes predictions of low-ranked classes to the same level since $\beta$ is constant and rank-irrelevant, which is conceptually sub-optimal. 

Our Zipf's loss is rank-relevant compared with label smoothing. In non-target classes, it encourages the high-ranked classes to keep larger predictions than low-ranked ones. Zipf's law distribution shows empirical success in our experiments against other rank-relevant distributions such as linear decay, more details are shown in Section \ref{sec:ablation study}.

%% file: experiment.tex
\begin{table}[!htp]\centering
\caption{Top-1 accuracy (\%) on CIFAR100, TinyImageNet image classification tasks with various model architectures. We report the mean and standard deviation over five runs with different random seeds. Vanilla indicates baseline results from the standard cross-entropy, the best results are indicated in bold, and the second-best results are indicated by underline. The performances of state-of-the-art methods are reported for comparison}\label{tab: 2} 
\scriptsize
\begin{tabular}{lcccc}\toprule
\multirow{2}{*}{Method} &CIFAR100 &TinyImageNet &CIFAR100 &TinyImageNet \\\cmidrule{2-5}
&\multicolumn{2}{c}{DenseNet121} &\multicolumn{2}{c}{ResNet18} \\\cmidrule{1-5}
Vanilla &77.86±0.26 &60.31±0.36 &75.51±0.28 &56.41±0.20 \\\cmidrule{1-5}
BAN~\cite{furlanello2018born} &78.39±0.14 &59.34±0.60 &76.96±0.04 &\underline{58.65±0.83} \\
BYOT~\cite{zhang2019your} &\underline{78.93±0.05} &60.54±0.02 &77.15±0.03 &57.84±0.15 \\
PS-KD~\cite{kim2021self} &78.82±0.10 &61.64±0.12 &76.74±0.06 &58.22±0.17 \\
DDGSD~\cite{xu2019data} &78.18±0.02 &60.80±0.30 &76.48±0.13 &58.52±0.12 \\
CS-KD~\cite{yun2020regularizing} &78.31±0.49 &\underline{62.04±0.09} &\textbf{78.01±0.13} &58.38±0.38 \\
LS~\cite{muller2019does} &78.12±0.45 &61.25±0.18 &77.31±0.28 &56.89±0.16 \\
TF-KD~\cite{yuan2020revisiting} &77.68±0.21 &60.17±0.57 &77.29±0.15 &56.67±0.05 \\\cmidrule{1-5}
Zipf's LS &\textbf{79.03±0.32} &\textbf{62.64±0.30} &\underline{77.38±0.32} &\textbf{59.25±0.20} \\
\bottomrule
\end{tabular}
\end{table}

\subsection{Experimental Detail}
\noindent \textbf{Datasets.} We conduct experiments in various image classification tasks to demonstrate our method's effectiveness and universality. Specifically, we use CIFAR100~\cite{Krizhevsky_2009_17719} and TinyImageNet\footnote{\url{https://www.kaggle.com/c/tiny-imagenet}} for small-scale classification tasks, and ImageNet~\cite{deng2009imagenet} for large-scale classification task. 
We also verify fine-grained classification performance with INAT21~\cite{van2021benchmarking} using the ``mini" training dataset.

\noindent \textbf{Training setups.}\label{train_setups} We followed the setups in recent related works~\cite{yun2020regularizing, chen2020online, guo2020online, yuan2020revisiting} and the popular open-source work\footnote{\url{https://github.com/facebookresearch/pycls}}. All experiments use MSRA initialization~\cite{he2015delving}, SGD optimizer with 0.9 momentum, 0.1 initial learning rate, 1e-4 weight decay, and standard augmentations including random cropping, and flipping. For small-scale CIFAR100 and TinyImageNet datasets, we use 32x32 resized input images, 128 batch size, and step learning rate policy which decreased to 1/10 of its previous value at 100th and 150th in the overall 200 epochs. All small-scale experiments are trained with single GPU. For large-scale ImageNet and INAT21 datasets, we use 224x224 resized input images, 256 batch size, and step learning rate policy which decreased to 1/10 of its previous value at 30th, 60th, and 90th in the overall 100 epochs. All large-scale experiments are trained with 4 GPUs.  

\noindent \textbf{Hyper parameters}. Our method has two hyperparameters in general, $\lambda$ and $\alpha$. $\lambda$ controls regularization strength and $\alpha$ controls the decay shape of Zipf's distribution, which is set to 1.0 in all experiments.
$\beta$ is only recommended for small resolution datasets such as CIFAR100 and TinyImageNet, to exploit the higher resolution intermediate feature map and make a more reliable ranking. A detailed hyperparameters ablation study is shown in the supplement. 

\begin{table}[!htp]\centering
\caption{Top-1 accuracy (\%) comparison with  state-of-the-art works. Experiments conduct on ImageNet, INAT21 image classification tasks with ResNet50, and on CIFAR100, TinyImageNet with DenseNet121.}\label{tab: 4} 
\scriptsize
\begin{tabular}{lccccc}\toprule
Method &CIFAR100 &TinyImageNet &ImageNet &INAT21 \\\cmidrule{1-5}
Vanilla &77.86 &60.31 &76.48 &62.43 \\\cmidrule{1-5}
CS-KD~\cite{yun2020regularizing} &\underline{78.31} &\underline{62.04} &\underline{76.78} &\underline{65.45} \\
LS~\cite{muller2019does} &78.12 &61.25 &76.67 &65.16 \\
TF-KD~\cite{yuan2020revisiting} &77.68 &60.17 &76.56 &62.61 \\\cmidrule{1-5}
Zipf's LS &\textbf{79.03} &\textbf{62.64} &\textbf{77.25} &\textbf{66.04} \\
\bottomrule
\end{tabular}
\end{table}

\subsection{General Image Classification Tasks}
First, we conduct experiments on CIFAR100 dataset and TinyImageNet dataset to compare with other related state-of-the-art methods, including self-knowledge distillation methods (BAN~\cite{furlanello2018born}), online knowledge distillation methods (DDGSD~\cite{xu2019data}, CS-KD~\cite{yun2020regularizing}), and label smoothing regularization method (label smoothing~\cite{muller2019does}, TF-KD~\cite{yuan2020revisiting}). Table \ref{tab: 2} shows the classification results of each method based on different network architectures. 
All experiments of the above methods keep the same setups for a fair comparison, details can be seen in \ref{train_setups} \textbf{Training setups}. For other hyper-parameters, we keep their original settings.\\

\noindent \textbf{Comparison with two-stage knowledge distillation.}
Two-stage knowledge distillation methods improve model accuracy using their previous models' dark knowledge. These methods rely on a pretrained model, which means they take twice or more training time than our method. We make a big advantage beyond BAN~\cite{furlanello2018born} (one step) as Table \ref{tab: 2} shows, specifically, Zipf's LS surpasses BAN~\cite{furlanello2018born} by 0.64\% and 3.28\% respectively on CIFAR100 and TinyImageNet based on DenseNet121. 


\noindent \textbf{Comparison with self-distillation.} As one type self-distillation utilizing contrastive samples, DDGSD and CS-KD realize respectively exploiting instance-wise and class-wise consistency regularization techniques. The data process of DDGSD or the pair sample strategy of CS-KD brings double iterations per training epoch. As shown in Table \ref{tab: 2}, Zipf's LS achieves 0.9\% and 0.73\% gains compared with DDGSD~\cite{xu2019data} respectively on CIFAR100 and TinyImageNet based on ResNet18 without more training iterations. As another type self-distillation with auxiliary parameters, BYOT~\cite{zhang2019your} squeeze deeper knowledge into lower network. Zipf's LS surpasses BYOT obviously on TinyImageNet.

\noindent \textbf{Comparison with label smoothing regularization.} Label smoothing~\cite{muller2019does} is a general effective regularizing method with manually designed soft targets. In Table \ref{tab: 2}, Zipf's LS beats label smoothing by 2.36\% and 1.39\% advantage on TinyImageNet respectively based on ResNet18 and DenseNet121. 

\begin{table}[!htp]\centering
\caption{Top-1 accuracy (gain) (\%) on ImageNet and INAT21 image classification tasks with various model architectures. Vanilla indicates baseline results from the cross-entropy, and the best results are indicated in bold}\label{tab: 3} 
\scriptsize
\begin{tabular}{lcccc}\toprule
Architecture &Method &ImageNet &INAT21 \\\cmidrule{1-4}
\multirow{3}{*}{ResNet18} &Vanilla &70.47 &54.31 \\
&Label Smooth &70.53(+0.06) &55.17(+0.86) \\
&Zipf's LS &\textbf{70.73 (+0.26)} &\textbf{56.36 (+2.03)} \\\cmidrule{1-4}
\multirow{3}{*}{ResNet50} &Vanilla &76.48 &62.43 \\
&Label Smooth &76.67(+0.19) &65.16(+2.73) \\
&Zipf's LS &\textbf{77.25 (+0.77)} &\textbf{66.04 (+3.61)} \\\cmidrule{1-4}
\multirow{3}{*}{ResNet101} &Vanilla &77.83 &65.60 \\
&Label Smooth &78.12(+0.29) &67.14(+1.54) \\
&Zipf's LS &\textbf{78.58 (+0.75)} &\textbf{68.45 (+2.85)} \\\cmidrule{1-4}
\multirow{3}{*}{ResNeXt50\_32x4d} &Vanilla &77.54 &66.36 \\
&Label Smooth &77.72(+0.18) &67.51(+1.15) \\
&Zipf's LS &\textbf{78.07 (+0.53)} &\textbf{69.24 (+2.88)} \\\cmidrule{1-4}
\multirow{3}{*}{ResNeXt101\_32x8d} &Vanilla &79.51 &70.35 \\
&Label Smooth &79.69(+0.18) &71.52(+1.17) \\
&Zipf's LS &\textbf{80.03 (+0.52)} &\textbf{72.18 (+1.83)} \\\cmidrule{1-4}
\multirow{3}{*}{DenseNet121} &Vanilla &75.56 &63.75 \\
&Label Smooth &75.59(+0.03) &64.60(+0.85) \\
&Zipf's LS &\textbf{75.98 (+0.42)} &\textbf{66.60 (+2.85)} \\\cmidrule{1-4}
\multirow{2}{*}{MobileNetV2} &Vanilla &65.52 &55.75 \\
&Label Smooth &65.71(+0.19) &56.29(+0.54) \\
&Zipf's LS &\textbf{66.03 (+0.51)} &\textbf{56.45 (+0.70)} \\
\bottomrule
\end{tabular}
\end{table}

\subsection{Large-scale and Fine-grained Image Classification Tasks}

\noindent \textbf{Comparison with state-of-the-art methods.} Our method is one-pass with almost zero extra computational or memory cost. Label smoothing and TF-KD are the two most related works with ours as shown in Fig. \ref{fig:intro_compare} and Table \ref{tab: method_compare }. And CS-KD is the most superior method on small-scale datasets besides ours as shown in Table \ref{tab: 2}. So we further compare our method with CS-KD, label smoothing and TF-KD on large-scale and fine-grained datasets. As shown in Table \ref{tab: 4}, our method shows much more superior performance while label smoothing and CS-KD methods have already improved baseline with significant margins. For instance, we surpass the second-best method by 0.47\% and 0.59\% respectively on ImageNet and INAT21 based on ResNet50.


\noindent \textbf{Improvements on various architectures.} We evaluate our method on various network architectures on ImageNet and INAT21. Not only the widely used families of ResNet~\cite{he2015deep} and ResNeXt~\cite{xie2017aggregated} are considered, but also the lighter architectures (such as MobileNetV2~\cite{sandler2019mobilenetv2}) are evaluated. Table \ref{tab: 3} shows our significant improvements compared with vanilla cross-entropy training on various network architectures based on ImageNet and INAT21 datasets. For instance, our method boosts baseline by 0.75\% and 2.85\% respectively on ImageNet and INAT21 with ResNet101.

\setlength{\tabcolsep}{6pt}
\begin{table}[!htp]\centering
\caption{Dense classification ranking v.s. logits-based ranking on TinyImageNet with ResNet18. CE indicates the standard cross-entropy loss. LR indicates that using the logits ranks. Dense1 indicates that using the last dense feature map of the last stage. Dense2 indicates that using the last dense feature map of the penultimate stage } \label{tab: 5}
\scriptsize
\begin{tabular}{lccccc}\toprule
\textcolor{black}{Method} &\textbf{CE} &\textbf{LR} &\textbf{Dense1} &\textbf{Dense2} &\textbf{Top-1 acc(\%)(gain)}  \\\midrule

\textcolor{black}{Vanilla} &\cmark & & & &\color{black}56.41 \\
\textcolor{black}{Logits-based} &\cmark &\cmark & &  &\color{black}(+1.70) \\
\textcolor{black}{Voting-based} &\cmark &\cmark &\cmark &  &\color{black}(+2.40) \\
\textcolor{black}{Voting-based} &\cmark &\cmark &\cmark &\cmark &\textbf{(+2.84)} \\
\bottomrule

\end{tabular}
\end{table}

\subsection{Ablation Study}\label{section:ablation}
\label{sec:ablation study}
\noindent \textbf{Dense classification ranking v.s. Logits-based ranking.}
We introduced two ranking metrics in section \ref{spatial-dense vote rank}, logits-based ranking and dense classification ranking. While logits-based ranking indeed boosts performance compared to the baseline (1.7\% as shown in the second row of Table~\ref{tab: 5}), we still find our strategy, the dense classification ranking, is necessary for optimal performance. When we replace the rank by the dense votes from the last few feature maps, the accuracy improvement goes to as high as 2.84\% (last row in Table~\ref{tab: 5}).

\noindent \textbf{Comparison of different distributions.} We constraint the logits rank or dense vote rank to obey Zipf's law due to the discovery that the output of deep neural network trained for a classification task follows the law as well. To demonstrate Zipf's priority, we conduct constant style, random style and decay style distributions on various datasets. As shown in Table \ref{tab: 6}, Zipf makes the best performance among these distributions. It's worth noting that, although inferior to Zipf's distribution, constant distribution also achieves satisfying performance which benefits from regularizing only non-target classes different from normal label smoothing. We speculate that label smoothing coupled with the target class that might hurt performance.

\begin{figure*}[t]
  \centering
\includegraphics[width=1.\textwidth]{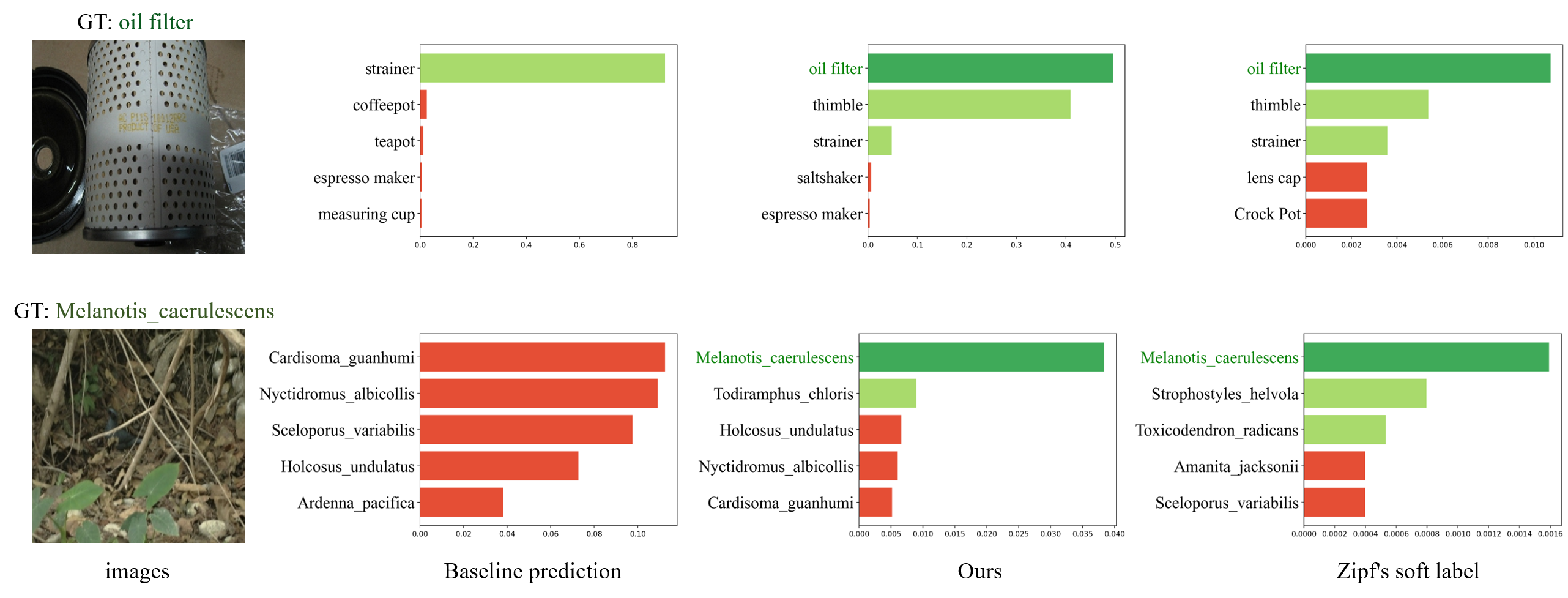}
  \caption{Comparison of top-5 predictions visualization. 
  The dark green, light green, and red colors denote ground-truth, similar and irrelevant categories respectively. More informative supervision and prediction are gained since categories that are more similar to ground truth arise top in Zipf's soft label (thimble and strainer vs oil filter).
  }
  \label{fig:result_sample}
\end{figure*}

%% file: discussion.tex


\noindent \textbf{Zipf's soft labels results in more reasonable predictions.} Fig.~\ref{fig:result_sample} illustrates the top-5 predictions of our proposed method compared with the baseline method. The images are sampled from ImageNet and INAT21 respectively. Our method makes more reasonable predictions. Not only the top-1 prediction is correct, but also more similar concepts arise in the top-5 prediction. This results from more informative soft labels from Zipf's LS, which provides meaningful representations as knowledge distillation for the network to better grasp the concept of similar categories.
\begin{figure*}[t]
  \centering
\includegraphics[width=1.\textwidth]{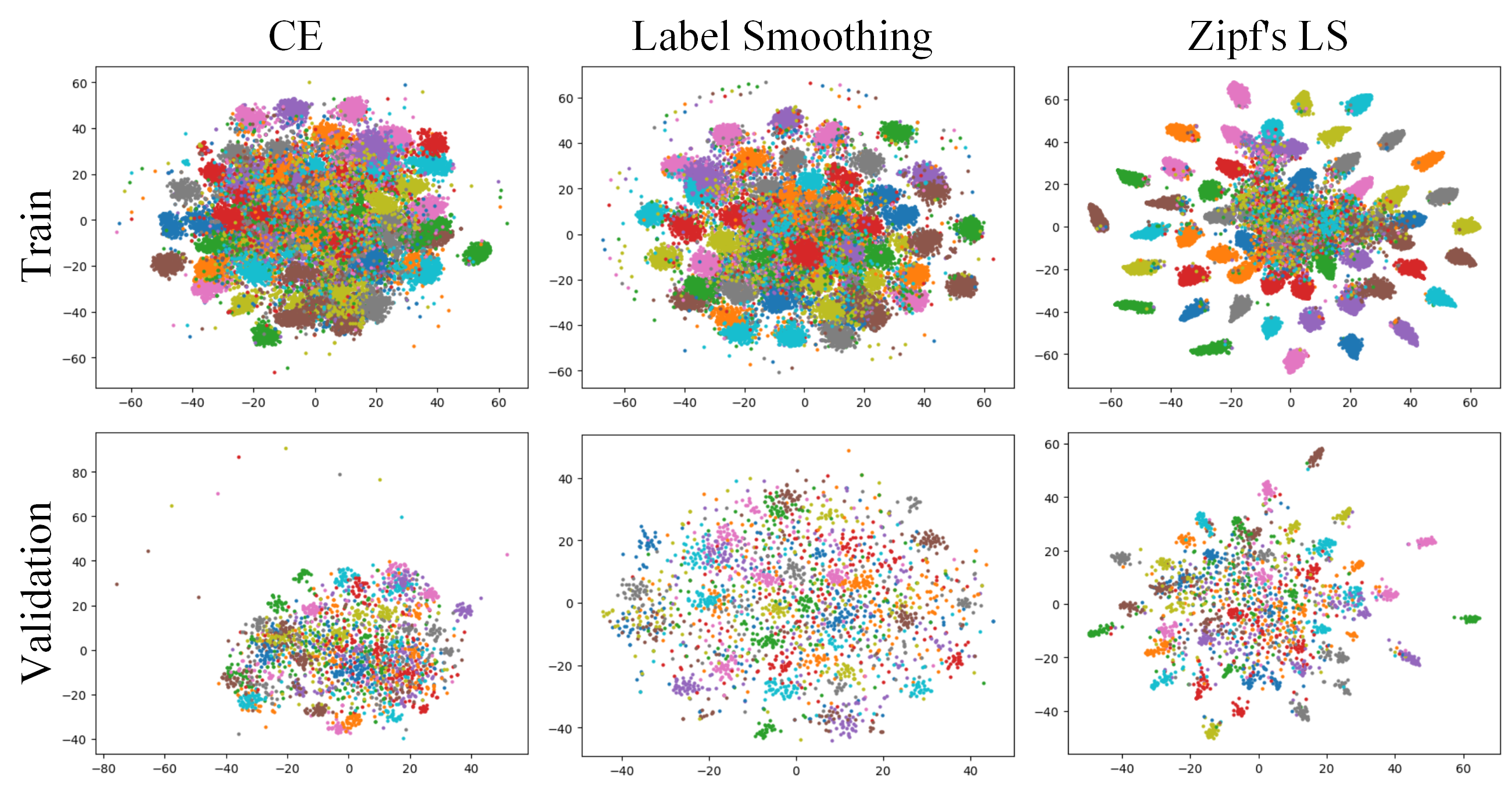}
  \caption{T-SNE~\cite{van2008visualizing} visualization on 50 random sampled classes of TinyImageNet for CE, Label Smoothing and Zipf's label smoothing.}
  \label{fig:feature_vis}
\end{figure*}

\setlength{\tabcolsep}{4pt}
\begin{table}[!htp]\centering
\caption{Comparison of the top-1 accuracy (\%) of different distributions and label smoothing. Experiments are conducted on TinyImageNet, ImageNet and INAT21. Zipf makes the best performance among these distributions}\label{tab: 6}
\scriptsize
\begin{tabular}{lcccc}\toprule
\textbf{Distribution} &\textbf{TinyImageNet} &\textbf{ImageNet} &\textbf{INAT21} \\\cmidrule{1-4}
Vanilla &56.41 &76.48 &62.43 \\ \cmidrule{1-4}
LS &56.89 &76.67 &65.16 \\
Constant &58.76 &77.09 &65.86 \\
Random\_Uniform &58.24 &76.89 &65.61\\
Random\_Pareto &58.52 &76.61 &65.9\\
Linear\_Decay &58.39 &76.87 &65.86\\\cmidrule{1-4}
Zipf &\textbf{59.25} &\textbf{77.25} &\textbf{66.04}\\ 
\bottomrule
\end{tabular}
\end{table}
\setlength{\tabcolsep}{1.4pt}

\noindent \textbf{Zipf's Label Smoothing achieves better representation learning for generalization.} We compare our Zipf's LS techniques with  cross-entropy training and uniform label smoothing training on the TinyImageNet dataset. As shown in Fig.~\ref{fig:feature_vis}, the intra-class distance in feature space learned from Zipf's label smoothing is more compact and the inter-class distance is more separate. Zipf's Label Smoothing achieves better representation learning for generalization. More discussion for generalization is shown in supplements.

\begin{figure*}[t]
  \centering
\includegraphics[width=1\textwidth]{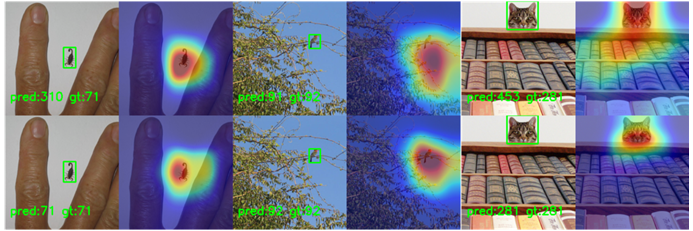}
  \caption{Comparison of small objects results between CE(above) and Zipf's LS(below).}
  \label{fig:small_objects}
\end{figure*}

\noindent \textbf{Non-target class dense classification ranking design.} To better rank the classes in the soft label, we exploit dense classification ranking instead of logits-based ranking. Thus, larger objects are favored and small object classification performance might degrade. In our design, the target class is excluded in $L_{Zipf}$ and included in $L_{CE}$ only, ensuring the target class provides the correct gradient regardless of the object size. To verify the small object performance, we collect the five smallest objects from each class\footnote{\url{https://image-net.org/data/bboxes_annotations.tar.gz}}, Zipf's LS still beat CE(69.25\% vs 67.72\%). The cam visualization for the small object case is shown in Fig \ref{fig:small_objects}


\noindent \textbf{Limitation.} Zipf's LS does not help in binary classification cases, since the $L_{zipf}$ only considers non-target class and would always be zero. Further, as mentioned in the method section, we exploit dense classification ranking to get more reliable ranking information than logit-based ranking, which is only available in image data but not in modals such as speech and language. To make Zipf's LS work in multi-modal data is considered as future work.




%% file: conclusion.tex
In this work, we propose an efficient and effective one pass self-distillation method named Zipf's Label Smoothing, which not only generates soft-label supervision in a teacher-free manner as efficient as label smoothing but also generates non-uniform ones as informative as more expensive self distillation approaches. Zipf's Label Smoothing consistently performs better than the uniform label smoothing method and other parameters-free one pass self-distillation methods, it could be one of the plug-and-play self-distillation techniques in your deep learning toolbox.